\documentclass{article}

\usepackage[final]{neurips_2023}

\usepackage[utf8]{inputenc} 
\usepackage[T1]{fontenc}    
\usepackage{hyperref}       
\usepackage{url}            
\usepackage{booktabs}       
\usepackage{amsfonts}       
\usepackage{nicefrac}       
\usepackage{microtype}      
\usepackage{xcolor}         
\usepackage{subcaption}
\usepackage{graphicx}
\usepackage{amsmath}

\usepackage{xcolor}

\title{Curriculum Learning 
for Cooperation in Multi-Agent Reinforcement Learning}

\author{
  Rupali Bhati \\
  Mila \& Laval University\\
  AI Redefined\\
  \texttt{rupali@ai-r.com} \\
  \And
  Sai Krishna Gottipati \\
  AI Redefined \\
  \texttt{sai@ai-r.com} \\
  \AND
  Clodéric Mars \\
  AI Redefined \\
  \texttt{cloderic@ai-r.com} \\
  \And
  Matthew E. Taylor \\
  University of Alberta \\ 
  AI Redefined \\
  \texttt{matt@ai-r.com} \\
}

\begin{document}

\maketitle

\begin{abstract}

While there has been significant progress in curriculum learning and continuous learning for training agents to generalize across a wide variety of environments in the context of single-agent reinforcement learning, it is unclear if these algorithms would still be valid in a multi-agent setting. 
In a competitive setting, a learning agent can be trained by making it compete with a curriculum of increasingly skilled opponents. 
However, a general intelligent agent should also be able to learn to act around other 
agents and cooperate with them to achieve common goals. 
When cooperating with other agents, the learning agent must (a) learn how to perform the task (or subtask), and (b) increase the overall team reward.
In this paper, we aim to answer the question of what kind of cooperative teammate, and a curriculum of teammates should a learning agent be trained 
with to achieve these two objectives. Our results on the game Overcooked show that a pre-trained teammate who is less skilled is the best teammate for overall team reward but the worst for the learning of the agent.
Moreover, somewhat surprisingly, a curriculum of teammates with \emph{decreasing} skill levels performs better than other types of curricula. 

\end{abstract}

\section{Introduction}
\label{Introduction}

Training agents that can generalize to previously unseen environments or scenarios has been a long-standing challenge in the field of reinforcement learning (RL). One approach to solving this problem is training an agent on a curriculum of increasingly complex tasks.
However, most curriculum learning work in RL is focused on training a single agent with increasingly complex environments
(\citep{UED, PLR}). In general, agents should be safe, efficient, and useful in a real-world context, where they often have to interact with other agents or humans of varying degrees of competence in the task.
In a competitive setting, one can train an agent by exposing the agent to a curriculum of opponents with increasing 
skill level(\citep{selfplay-barber, MAESTRO}). However, when in the cooperative setting, if a novice learning agent is cooperating with non-learning agents, it is not trivial to decide how a non-learning teammate should be selected to form a curriculum of teammates that would lead to efficient training of the learning agent.
This paper aims to frame and better understand this important problem.

In a cooperative setting, what does it mean to be a ``good teammate''? 
When cooperating with a novice learning agent, one could define a good teammate as someone who
can help the team achieve the most rewards, or 
as someone
who can help the novice agent learn to contribute to the shared task, or 
even some combination of these objectives.
The experiments in Section~\ref{sec:Experiment_Setup_and_Results} will demonstrate that these factors may not be directly correlated with each other and that presents a challenge on which factors to optimize for. 
To train a generally capable cooperative agent,
it would be ideal to train with \emph{all} possible teammates, which is intractable.
Therefore, we take advantage of curriculum learning to efficiently train generally capable agents through a curriculum of teammates.

In this work, we examine the ability of a novice learning agent to learn to cooperate with other agents of different skill levels in a multi-agent cooperative setting. We call the learning agent the \textit{student} and the agent with different skill levels that the student must cooperate with the \textit{teammate}. With the help of our experiments, consider the following questions: 
\begin{itemize}
    \item What makes a good teammate for a novice learning student?
    \item How can we construct a curriculum of teammates? 
    \item How do different curricula affect the student's learning?
\end{itemize}

We explore these questions empirically in the popular cooperative game Overcooked \citep{overcooked}. In Overcooked, two players cooperate to deliver as many onion soups as possible before the timer runs out. Before delivery, onions need to be placed in a pot, cooked, and plated. In our setting, since we have a student agent that can be paired with several types of teammates, that would require us to use an environment with heterogeneous agents. Since Overcooked does not pre-assign specific roles to either player and lets any player perform any of the sub-tasks, a very good player can account for the limitation of their teammate and ``do everything'', which makes the environment well suited to experiment with heterogeneous teams. 
In this paper, we analyze different pairings of student and teammate agents. Specifically, we pair a novice student agent with a low-skilled, medium-skilled, and highly-skilled teammate. Moreover, we train a novice student agent with different curricula of teammates.
We believe that the empirical analysis provided in this work serves as a foundation for further research on designing efficient curricula in multi-agent cooperative settings.

\section{Related Work}
\label{sec:Related_Work}

Beyond the seminal works outlined in \cite{2020JMLR} on curriculum learning, the generalization capabilities of agents were improved using different techniques including the framework of unsupervised environment design (UED)~\citep{UED}, genetic algorithms~\citep{poet} and domain randomization~\citep{domain-random}. Several works built on top of UED like Prioritized Level Replay~\citep{PLR} explore designing an efficient curricula of increasingly difficult environments for the learning agent to train on. In this paper, we are interested in the curricula of teammates in a multi-agent setting.

Curriculum learning has also been successfully applied in a multi-agent setting. Self-play \citep{selfplay-barber} is an approach for automatic curriculum generation in competitive, zero-sum games where an agent plays against past versions of itself in order to automatically generate a learning curriculum. However, this approach is only applicable to symmetric zero-sum games and does not lend itself to learning curricula for cooperative setups. \cite{MAESTRO} explores the curriculum over both the environment and the opponent, once again, in the zero-sum games setting.

\cite{Grupen2021Jun} present a method that uses 
environment-specific parameters to define the curriculum of the skill level of an agent combined with a curriculum of behavior that begins with single-agent learning before using the agent as a cooperative teammate. As they demonstrate their use on the pursuit-evasion task, they use velocity as a tool to determine the skill level of the multi-agent team which cannot be generalized to our setting. Moreover, their curriculum-driven DDPG strategy can be used only in the case of continuous action space, which is not the case in our setting.

\section{Background}
\label{sec:Background}

A standard Markov Decision Process (MDP) formulation is used for training reinforcement learning agents. 
In this paper, the agent(s) are trained in a multi-agent setting. They can be formulated within the framework of Multiagent MDP (MMDP)~\citep{MMDP}. 
As summarized in \cite{marl-mdp} and adapted to our notation, it can be defined with a tuple:

$$
\left\langle N, \mathbb{S},\left\{\mathbb{A}^i\right\}_{i \in\{1, \ldots, N\}}, P,\left\{R^i\right\}_{i \in\{1, \ldots, N\}}, \gamma\right\rangle
$$

where, $N$ is the number of agents,
$\mathbb{S}$ is the set of environmental states shared by all agents.
$\mathbb{A}^i$ is the set of possible actions of agent $i$. We denote $\mathbb{A}:=\mathbb{A}^1 \times \cdots \times \mathbb{A}^N$ as the joint action space.
$P:  \mathbb{S} \times \mathbb{A} \rightarrow \Delta(\mathbb{S})$ 
is the transition probability function that for each time step $t \in \mathbb{N}$, given agents' joint actions $\boldsymbol{a} \in \mathbb{A}$ and state $s_t \in \mathbb{S}$ gives the successor state $s_{t+1} \in \mathbb{S}$ at the next time step.
$R^i: \mathbb{S} \times \mathbb{A} \times \mathbb{S} \rightarrow \mathbb{R}$ is the reward function that returns a scalar value to the $i^{th}$ agent for a transition from $(s, \boldsymbol{a})$ to $s^{\prime}$. Lastly,
$\gamma \in[0,1]$ is the discount factor.

\section{Methods}
\label{sec:Methods}

\subsection{Creating a population of teammates}
\label{sec:Methods_Creating_a_population_of_teammates}

In order to answer the question of what makes a good teammate, we must first create a population of teammates to be able to compare them. We train two agents from scratch in the game of Overcooked~\citep{overcooked}. We randomly select one agent and call it the \textit{student} and the other agent
the \textit{teammate}. We denote the student policy by $\pi$ and the teammate policy by $\pi'$. We consider each agent in the environment to train independently and treat other agents as part of the environment. As the action space in our setting is discrete, the algorithm used to train these two agents is Independent DQN~\citep{iDQN}.
A typical DQN loss~\citep{dqn} at the $i^{th}$ iteration of training is given by: 
\begin{equation}
L_i\left(\theta_i\right)=\mathbb{E}_{s, a, r, s^{\prime}}\left[\left(y_i^{D Q N}-Q\left(s, a ; \theta_i\right)\right)^2\right]
\label{eq:dqnloss}
\end{equation}
where, 
$y_i^{D Q N}=r+\gamma \max _{a^{\prime}} Q\left(s^{\prime}, a^{\prime} ; \theta^{-}\right)$ is the temporal difference (TD) target, $\theta$ is the policy network and $\theta^{-}$ represents the parameters of a fixed target network. As we are in the Independent DQN setting, each agent has their own DQN network.

\paragraph{Population of teammates:} We denote teammate policies by $\pi'$ and the population of teammate policies by $\mathfrak{B}$. To create the population of teammates, we see it beneficial to have the members in the population have a diverse set of skill levels. The skill level of an agent is estimated by the duration for which the agent has been trained. We create a population of teammates by taking snapshots of the neural network of the teammate at different time intervals during the training $\{t_1, t_2, \dots t_K\}$ and denote these teammate policies by $\{ \pi'_{t_1}, \pi'_{t_2}, \dots  \pi'_{t_K} \}$, where $K$ is the maximum number of episodes that a teammate can be trained till.

\subsection{Training with teammates}
\label{sec:Method_Training_with_teammates}

\paragraph{Train with pre-trained agents}

To determine which teammate is a good teammate, we pair a new student agent that learns from scratch with a pre-trained (non-learning) teammate that is sampled from the population of teammates $\mathfrak{B}$. We specifically use three teammates $\{ \pi'_{t_1}, \pi'_{t_2}, \pi'_{t_3} \}$ where $t_1 < t_2 < t_3$. We call the $\pi'_{t_1}$ as the less trained (or less skilled) agent, the $\pi'_{t_2}$ as the medium trained agent, and the $\pi'_{t_3}$ as the highly trained agent.

\paragraph{Train with a curriculum of pre-trained agents}

In order to train with a curriculum of pre-trained agents, we can make several combinations of the three pre-trained agents in $\mathfrak{B}$ (i.e., low-skilled, medium-skilled, and highly-skilled). We divide the training procedure of the student agent into three equal parts of the number of episodes. For the total number of episodes is $K$, we pair the student agent with a different agent for every $K/3$ episodes. We create two curricula of teammates:

\begin{enumerate}
    \item \textbf{Increasing curriculum}: We increase the skill level of the teammates during training. Therefore, during the first $K/3$ episodes, the student agent is trained alongside the low-skilled teammate, then for the next $K/3$ episodes, it is trained alongside the medium-skilled teammate, and for the last $K/3$ episodes, it is trained alongside the highly-skilled teammate.
    \item \textbf{Decreasing curriculum}: We decrease the skill level of the teammates during training. Therefore, similar to the increasing curriculum, the student is first paired with the highly-skilled teammate, then the medium-skilled teammate, and finally the low-skilled teammate, each for $K/3$ episodes.
\end{enumerate}

\section{Experiment Setup and Results}
\label{sec:Experiment_Setup_and_Results}

The aim of our experiments are the following: (i) Find who a good teammate is from a pool of teammates, (ii) Validate whether training an agent with a curriculum of teammates is helpful in the cooperative setting, and (iii) What kind of curriculum of teammates meets different objectives of cooperation.

\subsection{Environment}

\begin{figure}[h]
    \centering
    \includegraphics[width=0.25\linewidth]{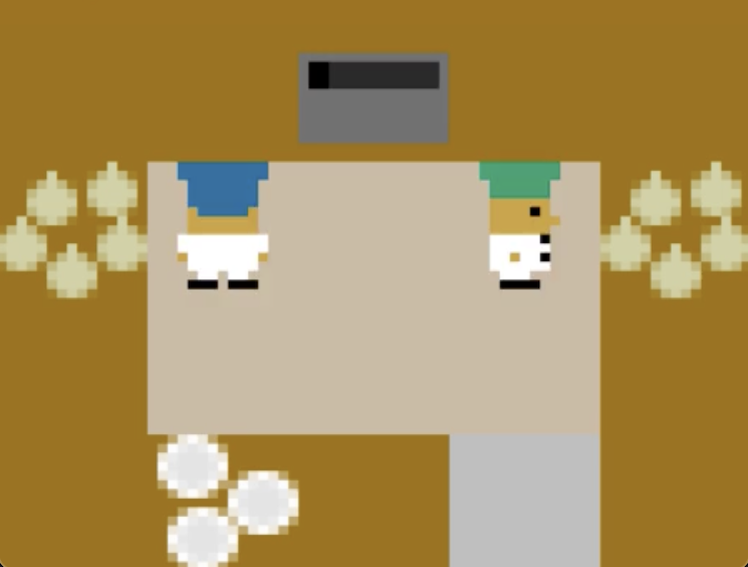}
    \caption{Overcooked Environment}
    \label{fig:environment}
\end{figure}

We use the environment Overcooked as shown in Figure~\ref{fig:environment}.
In this environment, there are two agents who may choose to cooperate or not. The overall task is to deliver as many fully cooked soups as possible within the time limit $\mathbb{N}$. The process to deliver a fully cooked soup involves several steps. First, an agent must pick up one onion at a time and place it on the stove. Once there are three onions on the stove, the soup starts cooking. The soup cooks for a certain amount of timesteps and the agents do not need to monitor it. Once cooked, an agent must pick up a bowl and pour the soup into the bowl. Lastly, an agent must deliver that cooked soup (along with the bowl) at the delivery counter. Upon delivering the soup, the agent receives a reward. 

The two agents can be trained using any combinations of individual rewards. For example, the agents can be trained using their individual rewards. This could motivate the agents to want to deliver the soup and perhaps not let the other agent deliver it. We use the cooperative setting, where both agents are rewarded equally and trained on the combined team reward.

\subsection{Creating the population of teammates} 
\begin{figure}[h]
   \centering
    \begin{subfigure}{0.32\textwidth}
        \centering
        \includegraphics[width=\linewidth]{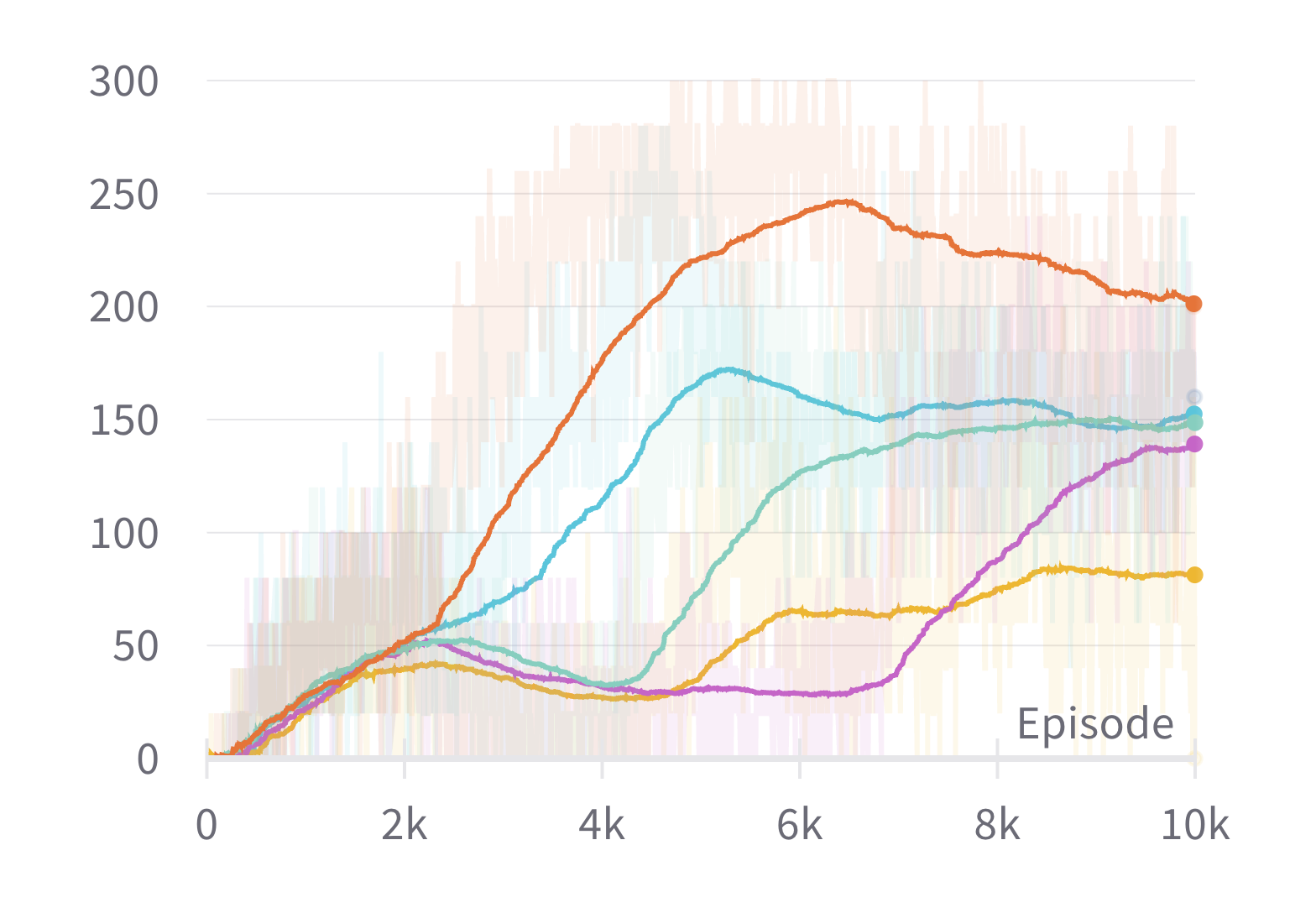}
        \caption{student reward}
        \label{fig:IDQN_same_reward_10K_sparse_reward_agent_1}
    \end{subfigure}
    \hfill
    \begin{subfigure}{0.32\textwidth}
        \centering
        \includegraphics[width=\linewidth]{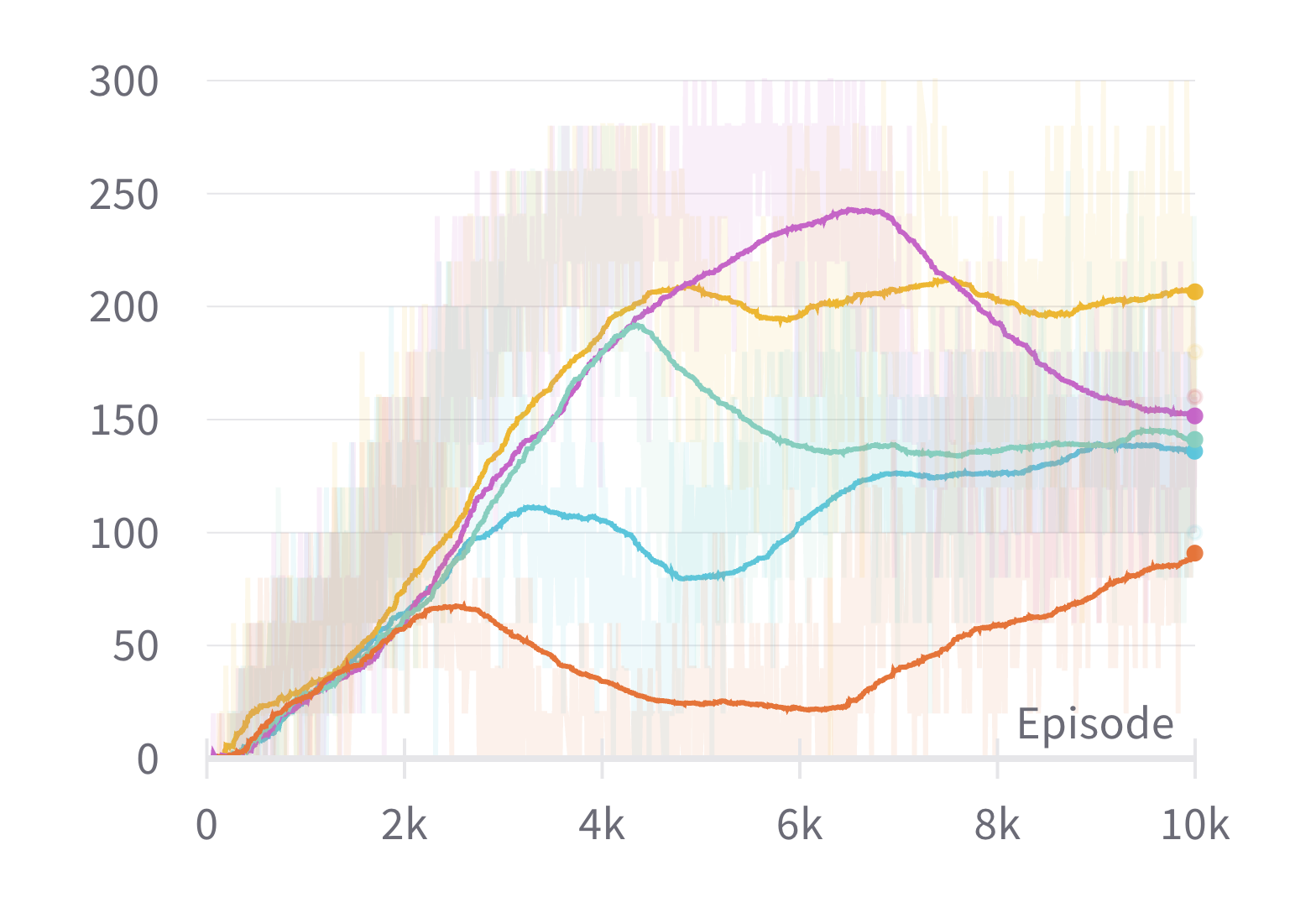}
        \caption{teammate reward}
        \label{fig:IDQN_same_reward_10K_sparse_reward_agent_2}
    \end{subfigure}
    \hfill
    \begin{subfigure}{0.325\textwidth}
        \centering
        \includegraphics[width=\linewidth]{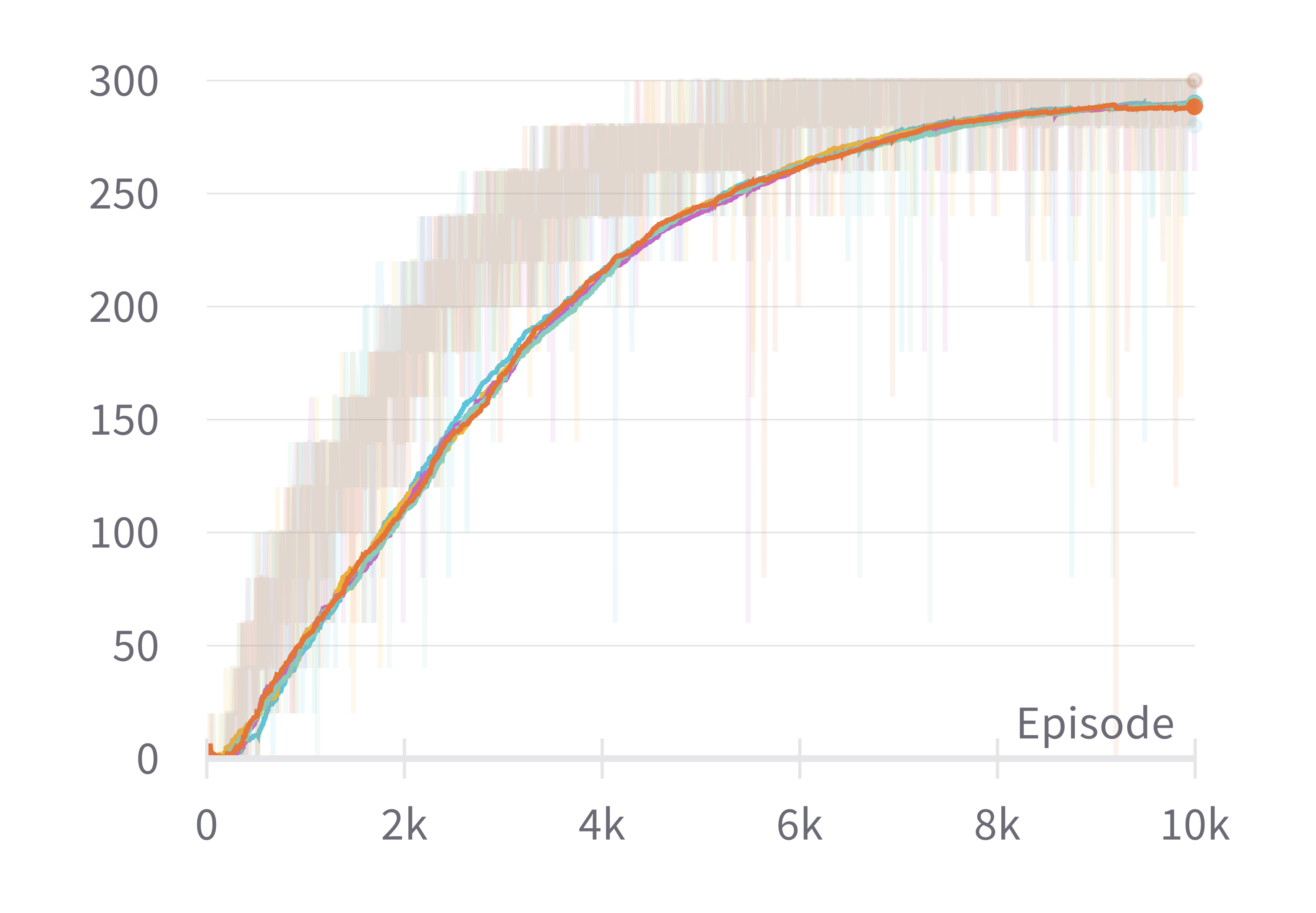}
        \caption{total reward}
        \label{fig:IDQN_same_reward_10K_total_sparse_reward}
    \end{subfigure}
\caption{Individual and total rewards for the student and teammate agents when both are trained together from scratch. All plots correspond to runs for 5 seeds.}    
\label{fig:IDQN_same_reward_10K}    
\end{figure}

As explained in Section~\ref{sec:Methods_Creating_a_population_of_teammates}, we first train two agents from scratch using Independent DQN. We train both agents together for $K=$10,000 episodes. We use epsilon decay for the policy where epsilon decays from $\epsilon_{start}$ and goes to $\epsilon_{end}$ at the end of the training. Figure~\ref{fig:IDQN_same_reward_10K_total_sparse_reward} shows the rolling average of the total team reward for 5 seeds of training.  Note that the value of the total reward converges after 10,000 episodes but we clip the rest of the graph for better comparison. The values of the hyperparameters can be found in Table~\ref{tab:Values_of_hyperparameters} in the Appendix.

In Figures~\ref{fig:IDQN_same_reward_10K_sparse_reward_agent_1} and ~\ref{fig:IDQN_same_reward_10K_sparse_reward_agent_2}, we
track the individual reward for both the student and the teammate. 
The variance across different seeds in the individual reward of the student and the teammate shows that some agents may learn to deliver the dish more often than other agents. Recall that both agents are trained using the total reward, therefore they are trained to be indifferent to who delivers the dish. It is interesting to note that different seeds learn a wide variety of behaviors when it comes to delivering the dish. 

We create the population of teammates $\mathfrak B$, using three snapshots during training at $t_1=$2000, $t_2=$5000, and $t_3=$10000 episodes making $\mathfrak B = \{ \pi'_{t_1}, \pi'_{t_2}, \pi'_{t_3} \} = \{ \pi'_{2K}, \pi'_{5K}, \pi'_{10K} \}$.

\subsection{Training with teammates}

\paragraph{Train with pre-trained agents} 

\begin{figure}[h]
        \centering
        \includegraphics[width=\linewidth]{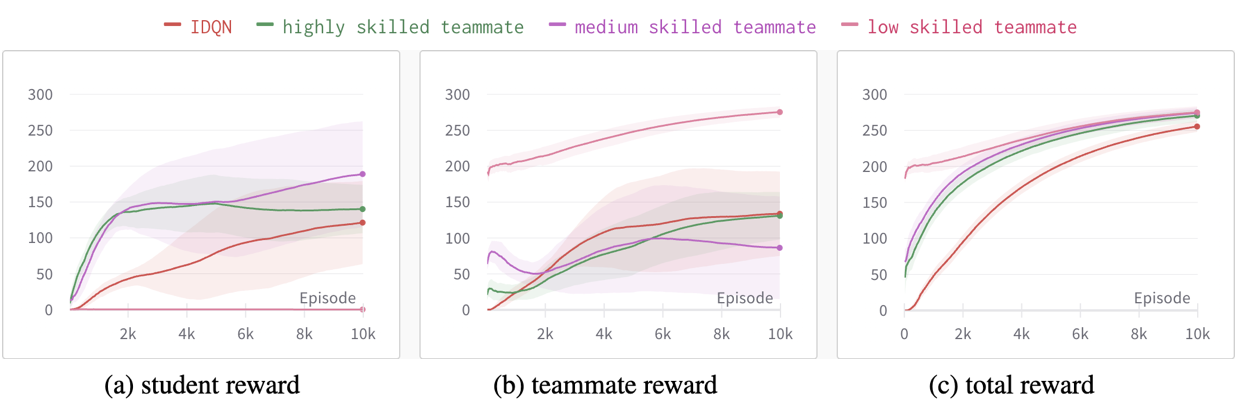}
\caption{Individual and total reward for the student and teammate agents when the student (that is trained from scratch) is paired with a non-learning teammate. All means and standard deviations are aggregated over 5 seeds.}    
\label{fig:IDQN_vs_Frozen_teammates}    
\end{figure}

We train a new student agent with different pre-trained teammates sampled from $\mathfrak B$. Note that the weights of the teammate's network are frozen meaning that the teammate is not learning, while the student agent is learning from scratch.
Figure~\ref{fig:IDQN_vs_Frozen_teammates} shows the mean and standard deviation of the individual and total rewards obtained for the team when a student agent is paired with a pre-trained teammate--either $\pi'_{2K}$ (low skilled), $\pi'_{5K}$ (medium skilled), or $\pi'_{10K}$ (highly skilled)--or with a teammate that learns from scratch (IDQN).

We observe that training with a pre-trained teammate is always better than training with a new agent from scratch for the total reward. This result is not surprising as it might be a result of the fact that any pre-trained teammate with non-updating weights in the neural network reduces the non-stationarity of the environment and hence leads to faster convergence to the higher reward. The interesting result here is that the low-skilled teammate not only starts at a higher total reward but also reaches the highest reward the fastest and stays the highest. This suggests that the low-skilled teammate is the best teammate for maximizing the total reward. Moreover, the rewards suggest that the order of teammates to maximize total reward is low skilled > medium skilled > highly skilled. 

On the flip side, the student reward is negligible when it is paired with the low-skilled teammate. This indicates that this pairing leads to the student agent to turn into a ``lazy agent", which is a common problem in cooperative multi-agent RL. The teammate that leads to the highest student reward while also achieving a high total reward at convergence is the medium-skilled teammate.

\paragraph{Train with a curriculum of pre-trained agents:} 

\begin{figure}[h]
   \centering
    \includegraphics[width=\linewidth]{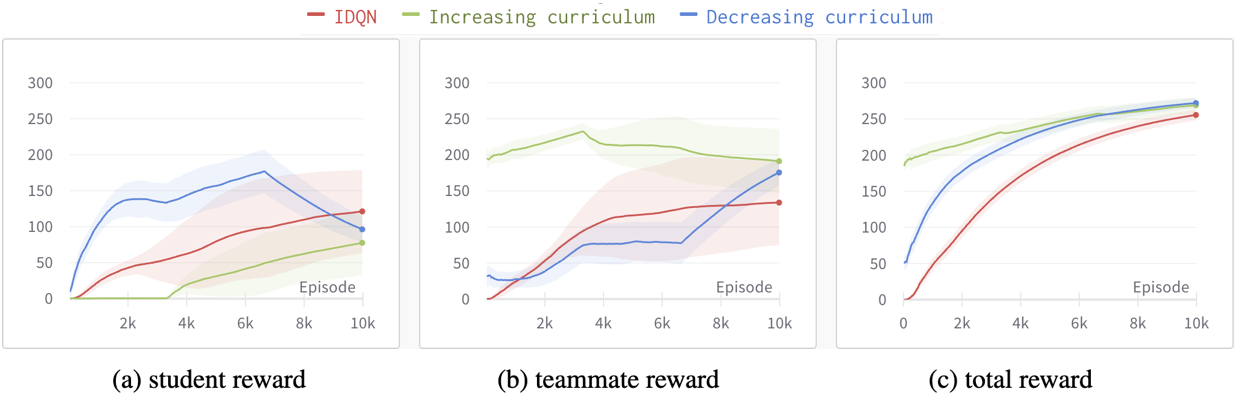}
\caption{Individual and total reward for the student and teammate agents when the student is trained from scratch with a curriculum of teammates whose weights are frozen. All means and standard deviations are aggregated over 5 seeds.}    
\label{fig:IDQN_vs_Curriculum_teammates}    
\end{figure}

As explained in Section~\ref{sec:Method_Training_with_teammates}, the student agent is trained with two types of curriculum of teammates: increasing and decreasing. In Figure~\ref{fig:IDQN_vs_Curriculum_teammates}, we show the mean and standard deviation of the individual and total rewards of the team where the student agent is trained with an increasing curriculum, a decreasing curriculum, and a teammate learning from scratch (IDQN). In the case of the increasing curriculum, the student agent is first paired with the $\pi'_{2K}$ teammate for the first 3333 episodes, then with the $\pi'_{5K}$ teammate for the next 3333 episodes, and finally with the $\pi'_{10K}$ teammate for the remaining 3334 episodes. Similarly, for the decreasing curriculum, every 3333 episodes, the teammate changes to a lower-skilled teammate.

Here, we observe that a curriculum of teammates is better than training with an agent who learns from scratch for total reward. 
Moreover, the increasing curriculum has a higher total reward at the beginning, however, the decreasing curriculum surpasses the reward of the increasing curriculum after 2/3rd of the training. Note that, this is in line with our observations of when training with the pre-trained agent and how the total reward is highest with the low-trained agent. In the decreasing curriculum, the teammate after 2/3rd of the training is the low-skilled agent which might explain the higher reward.

We can also observe that the decreasing curriculum encourages the teammate agent to receive a lower reward whereas the increasing curriculum teammate maintains a high individual reward for the teammate right from the beginning as shown in Figure~\ref{fig:IDQN_vs_Curriculum_teammates}(b). Ultimately we see in Figure~\ref{fig:IDQN_vs_Curriculum_teammates}(a), that the student agent when trained with the decreasing curriculum has a higher reward initially during training which it forgoes in the last $K/3$ episodes.

\subsection{Total reward and individual learning}

\begin{figure}[h]
   \centering
    \includegraphics[width=\linewidth]{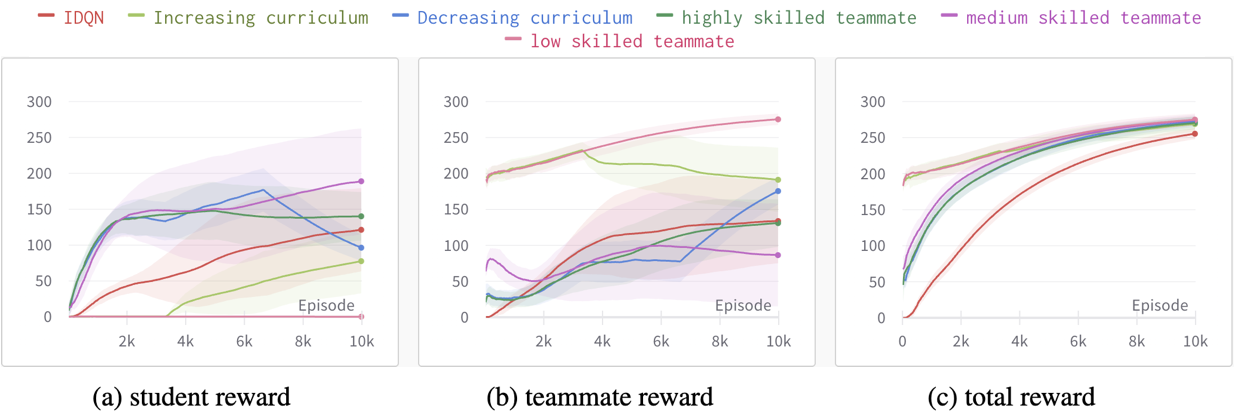}
\caption{Individual and team rewards for the student and teammate agents when the student is trained from scratch with different types of teammates. All means and standard deviations are aggregated over 5 seeds.}    
\label{fig:IDQN_vs_Frozen_vs_Curriculum_teammates}    
\end{figure}

In Figure~\ref{fig:IDQN_vs_Frozen_vs_Curriculum_teammates}, we compare the individual and total reward for all the runs including training in the Independent DQN setting, training with pre-trained teammates, and training with a curriculum of pre-trained teammates. In Figure~\ref{fig:IDQN_vs_Frozen_vs_Curriculum_teammates}(c), we see that near the end of the training, there are two training methods that reach the highest total team reward. The methods are the low-skilled teammate and medium-skilled teammate runs. Out of the low-skilled and the medium-skilled teammates, we see in Figure~\ref{fig:IDQN_vs_Frozen_vs_Curriculum_teammates}(b) that the medium-skilled teammate has much lower individual reward than the low-skilled teammate. This is also reflected in Figure~\ref{fig:IDQN_vs_Frozen_vs_Curriculum_teammates}(a), where the student agent has consistently zero reward when trained with the low-skilled teammate. This clearly shows that 
the medium-skilled teammate leads to (i) one of the highest team rewards at the end of the training, and (ii) the highest individual reward for the student.

\section{Conclusion/Discussion}
\label{sec:Conclusion}

Our work offers insight into what type of teammate is required in a cooperative setup. With the help of our experiments, we validate that the selection of this teammate is not trivial. Our results demonstrate that a pre-trained teammate who is less skilled (trained) is the best teammate for overall team reward but the worst for the learning of the student agent who is learning from scratch.
Moreover, we discover that a curriculum of pre-trained agents is indeed better than training an agent with another learning agent (that learns from scratch). When compared with other curricula, we see that a curriculum of teammates with decreasing skill levels is the optimal choice for both team reward as well as the learning of the agent. However, when comparing a curriculum of teammates with just a single pre-trained teammate, the optimal choice for both team reward, as well as the learning of the agent, is the medium-skilled teammate.

We hope to extend this work to other cooperative environments and see what kind of teammates are good in those settings and whether a curriculum can help with training in those environments. Moreover, with the preliminary work done in this paper, we hope to inspire future work in the direction of curriculum learning in the cooperative setting.

\newpage

\bibliographystyle{abbrvnat}

\newpage
\appendix

\section{Appendix}

\begin{table}[h]
  \caption{Values of hyperparameters}
  \label{tab:Values_of_hyperparameters}
  \centering
  \begin{tabular}{ll}
    \toprule
    Hyperparameter     & Value     \\
    \midrule
    Maximum number of steps in an episode $\mathbb{N}$ & 500   \\
    Number of episodes $K$     & 10,000     \\
    $t_1$   &   2,000    \\
    $t_2$   &   5,000    \\
    $t_3$   &   10,000    \\    
    Reward for delivering a dish &   20  \\
    $\epsilon_{start}$  & 0.9  \\ 
    $\epsilon_{end}$  & 0.05  \\ 
    Discount factor $\gamma$    &   0.99    \\
    \bottomrule
  \end{tabular}
\end{table}

\end{document}